%% file: main.tex
\documentclass{article}

\usepackage{arxiv}

\usepackage[utf8]{inputenc} 
\usepackage[T1]{fontenc}    
\usepackage{hyperref}       
\usepackage{url}            
\usepackage{booktabs}       
\usepackage{amsfonts}       
\usepackage{nicefrac}       
\usepackage{microtype}      
\usepackage{lipsum}		
\usepackage{graphicx}
\usepackage{natbib}
\usepackage{doi}
\usepackage{amsmath}
\usepackage{bbm}
\usepackage{placeins}
\usepackage{multirow}

\title{HiRef: Leveraging Hierarchical Ontology and Network Refinement for Robust Medication Recommendation}

\newcommand{\corrauth}{\thanks{Corresponding author.}}
\newcommand{\ourmodel}{HiRef}

\author{{Yan Ting Chok} \\
	Korea University\\
	Seoul, South Korea \\
	\texttt{yanting1412@korea.ac.kr} \\
	\And
        {Soyon Park} \\
	Korea University\\
	Seoul, South Korea \\
	\texttt{soyon\_park@korea.ac.kr} \\
	\And
        {Seungheun Baek} \\
	Korea University\\
	Seoul, South Korea \\
	\texttt{sheunbaek@korea.ac.kr} \\
	\And
        {Hajung Kim} \\
	Korea University\\
	Seoul, South Korea \\
	\texttt{hajungk@korea.ac.kr} \\
	\And
        Junhyun Lee\corrauth \\
	Korea University\\
	Seoul, South Korea \\
	\texttt{ljhyun33@korea.ac.kr} \\
	\And
        Jaewoo Kang\corrauth \\
	Korea University\\
	Seoul, South Korea \\
	\texttt{kangj@korea.ac.kr} \\
}

\renewcommand{\shorttitle}{\textit{arXiv} Template}

\hypersetup{
pdftitle={A template for the arxiv style},
pdfsubject={q-bio.NC, q-bio.QM},
pdfauthor={Yan Ting Chok},
pdfkeywords={Medication Recommendation, Medical Ontology, Network Refinement},
}

\begin{document}
\maketitle

\begin{abstract}
    Medication recommendation is a crucial task for assisting physicians in making timely decisions from longitudinal patient medical records. However, real-world EHR data present significant challenges due to the presence of rarely observed medical entities and incomplete records that may not fully capture the clinical ground truth. While data-driven models trained on longitudinal Electronic Health Records often achieve strong empirical performance, they struggle to generalize under missing or novel conditions, largely due to their reliance on observed co-occurrence patterns. To address these issues, we propose \textbf{Hierarchical Ontology and Network Refinement for Robust Medication Recommendation (\ourmodel)}—a unified framework that combines two complementary structures: (i) the hierarchical semantics encoded in curated medical ontologies, and (ii) refined co-occurrence patterns derived from real-world EHRs. We embed ontology entities in hyperbolic space, which naturally captures tree-like relationships and enables knowledge transfer through shared ancestors, thereby improving generalizability to unseen codes. To further improve robustness, we introduce a prior-guided sparse regularization scheme that refines the EHR co-occurrence graph by suppressing spurious edges while preserving clinically meaningful associations. Our model achieves strong performance on EHR benchmarks (MIMIC-III and MIMIC-IV) and maintains high accuracy under simulated unseen-code settings. Extensive experiments with comprehensive ablation studies demonstrate \ourmodel's resilience to unseen medical codes, supported by in-depth analyses of the learned sparsified graph structure and medical code embeddings.
\end{abstract}


\section{Introduction}
Medication recommendation is a cornerstone of clinical decision support. Faced with evolving standards of care and time constraints, clinicians must determine safe and effective therapies from fragmented, longitudinal Electronic Health Records (EHRs) \cite{all2019all,sutton2020overview,ali2023deep,mishra2024knowledge}. EHRs contain standardized diagnosis, procedure, and medication codes that document patient care over time. The availability of this structured clinical data has created opportunities for machine learning approaches to identify meaningful patterns and associations among these         entities. The increasing scale of EHR data, combined with advances in deep learning, has thus led to the widespread adoption of computational models for medication recommendation.

Prior works on medication recommendation has generally followed two directions. The first direction involves modeling pharmacological or molecular properties to improve the safety and efficacy of drug combinations, sometimes incorporating drug–drug interaction constraints during prediction \cite{yang2021safedrug,yang2023molerec}. The second direction involves learning directly from EHRs, capturing longitudinal patterns across visits and co-occurrences among diagnoses, procedures, and medications occasionally augmented with patient-centric knowledge graphs or distilled large language models \cite{shang2019gamenet,ijcai2019p825,kim2025hi,singhal2023large}. Although these approaches achieve strong in-distribution accuracy, they struggle to generalize under missing-information or unseen-code scenarios.

In practice, the frequency of medical codes in EHR data is highly imbalanced due to the natural rarity of certain medical conditions or prescriptions. While rare codes appear infrequently in training data, they often represent clinically significant conditions that require specific therapeutic interventions. When such codes appear at inference time, they can be critical for accurate medication recommendation. A robust medication recommender must therefore be able to handle these rare codes effectively, potentially by leveraging their relationships within medical ontologies to connect them to appropriate treatments. However, most existing models struggle with unseen codes because they rely heavily on observed co-occurrences, limiting their ability to represent codes absent during training \cite{song2021generalized}. Beyond code rarity, real-world EHRs suffer from various data quality issues that affect model performance. Incomplete records may arise from fragmented care across multiple institutions, inconsistent coding practices, or documentation gaps. Additionally, temporal changes such as evolving diagnostic criteria or drug substitutions during supply shortages can introduce inconsistencies in the data. Models trained on such imperfect data are prone to learning spurious associations, ultimately undermining their robustness in clinical deployment. 

We address challenges by integrating medical ontologies with EHR co-occurrence graphs. Our approach combines two complementary sources of structure: (i) the hierarchical semantics curated by medical ontologies, and (ii) refined co-occurrence patterns derived from real-world EHRs. These two sources provide distinct but complementary signals. Medical ontologies encode clinically meaningful semantic relationships—such as the distinction between Type I and Type II diabetes—that do not typically co-exist and is difficult to infer reliably from statistical patterns alone ~\cite{bagley2016constraints,breeyear2024development,elsayed20252}. In contrast, EHR-based co-occurrence graphs capture real-world treatment patterns and frequently associated conditions, such as the common co-occurrence of diabetes, obesity, and hypertension in patient records. By leveraging oncology structure to transfer information through shared ancestors and using EHR evidence to reflect real-world treatment regularities, our model can make more informed and generalizable drug recommendations. Furthermore, pruning spurious correlations in co-occurrence relationships reduces noise inherent in incomplete EHR data, thereby improving model robustness.

To this end, we propose \textbf{Hierarchical Ontology and Network Refinement for Robust Medication Recommendation (\ourmodel)}, a framework that unifies ontology-aware and data-driven representations for medication recommendation. 
First, we embed diagnosis, procedure, and medication ontologies in \emph{hyperbolic} space, a geometry well-suited for representing tree-like structures with minimal distortion. 
By aligning parent–child relationships and aggregating ancestor information via Möbius operations, the resulting embeddings enable knowledge transfer along the hierarchy. When an entity’s code exists but was unseen during training, its representation inherits informative signals from ancestors and siblings, facilitating zero- or few-shot generalization without retraining. 
Second, we construct a directed, cross-entity-type EHR co-occurrence graph from observational data and refine it using a prior-guided sparse regularization scheme. 
We initialize edges with visit-level conditional co-occurrence probabilities, then learn a \emph{sparsified} attention graph that suppresses spurious edges while preserving clinically essential ones. 
This produces compact neighborhoods that enhance model robustness, improve computational efficiency, and yield localized, inspectable rationales for predictions. 
Additionally, we implement a lightweight \emph{convex gating} module that adaptively fuses ontology-aware and co-occurrence pathways on a per-entity basis, by learning which source of evidence should dominate each recommendation.

\ourmodel~ targets two aspects that are important in clinical deployment.
i) \textbf{Generalizability} is achieved through hierarchical ontology encoding, which provides complementary information about the semantic meaning, properties, and usage of medical entities. This enables robust inference even when encountering unseen or rarely observed codes. ii) \textbf{Robustness} is ensured by a sparse, prior-guided co-occurrence graph encoder that resists noise in incomplete EHRs by discarding spurious correlations, thereby improving both the reliability and interpretability of the recommendations. 

We evaluate \ourmodel~ on public EHR benchmarks (MIMIC-III~\citep{johnson2016mimic} and MIMIC-IV~\citep{johnson2023mimic}), assessing performance in both in-distribution and unseen settings, the latter created by masking critical inputs during training. Across these settings, \ourmodel~ achieves strong in-distribution performance while maintaining accuracy in unseen-code scenarios. Ablation studies confirm that hierarchical ontology encoding effectively complements co-occurrence patterns, and that sparsity regularization in the graph encoder captures essential patterns that improve recommendation accuracy. We further analyze the learned EHR graph to interpret predictions and uncover meaningful, evidence-based associations among medical entities. Furthermore, we visualize the learned embeddings to verify that model structures medical codes in a semantically meaningful way.
Our contributions are as follows:
\begin{itemize}
\item We propose \ourmodel, a medication recommendation framework that unifies hierarchical ontology embeddings with a prior-guided, sparsified EHR co-occurrence graph. This design enables knowledge transfer from ancestors and siblings in the ontology to handle \emph{pre-existing but previously unseen or rarely observed} codes.
\item We incorporate sparsity regularization into the learning of EHR-based co-occurrence patterns, improving not only model robustness but also computational efficiency and interpretability.
\item We conduct extensive experiments on MIMIC-III and MIMIC-IV under unseen settings, along with comprehensive ablation studies. Our results demonstrate state-of-the-art accuracy and resilience to unseen medical codes, supported by in-depth analyses of the learned sparsified graph structure and medical code embeddings.
\end{itemize}

\section{Preliminaries}

\textbf{Electronic Health Records (EHR).} 
EHR data contains multimodal information, including structured information such as lab test results and prescription history, as well as unstructured data like clinical notes written by healthcare providers. Among them, diagnosis, procedure, and medication records are widely used for clinical predictive modeling due to their standardized format and clinical relevance. These codes provide a discrete and systematic representation of a patient's clinical status and treatment history, making them particularly suitable for tasks such as medication recommendation.

\textbf{Medical Code Systems.} All diagnosis, procedure, and medication entities are represented as medical codes, each structured within standardized coding systems. International Classification of Diseases, Ninth Revision (ICD-9)~\citep{hirsch2016icd} is the official system of assigning codes to diagnoses and procedures associated with hospital utilization in the US and is categorized in a structured manner.  For medications, the Anatomical Therapeutic Chemical (ATC)~\citep{world2021anatomical} classification system categorizes active substances into five levels based on the organ or system they act upon and their therapeutic, pharmacological, and chemical characteristics. In this work, we adopt ATC Level 3 for representing medications following prior studies. These medical ontologies provide a hierarchical structure that captures both semantic similarity and clinical relevance among medical codes, enabling to learn more meaningful relationships among medical entities. Moreover, it helps reduce overfitting to dataset-specific co-occurrence patterns, grounding the model in medically valid relationships, enhancing its transferability across settings. 

\textbf{Hyperbolic embedding.} Hyperbolic space is a non-Euclidean geometry characterized by constant negative curvature, where its volume grows exponentially with radius. This allows hyperbolic space to embed hierarchical or tree-like structures more efficiently and with lower distortion than Euclidean space. Among various models of hyperbolic space, the Poincaré ball model is one of the most widely used in representation learning. In this model, each point lies inside an open unit ball $\mathbb{B}^n = \left\{ \mathbf{x} \in \mathbb{R}^n : \|\mathbf{x}\| < 1 \right\}$. This property makes hyperbolic geometry particularly well-suited for representing symbolic data with inherent hierarchies, such as taxonomies, ontologies, or medical code systems. 
\begin{figure}[htbp]
  \centering
  \includegraphics[width=0.7\linewidth]{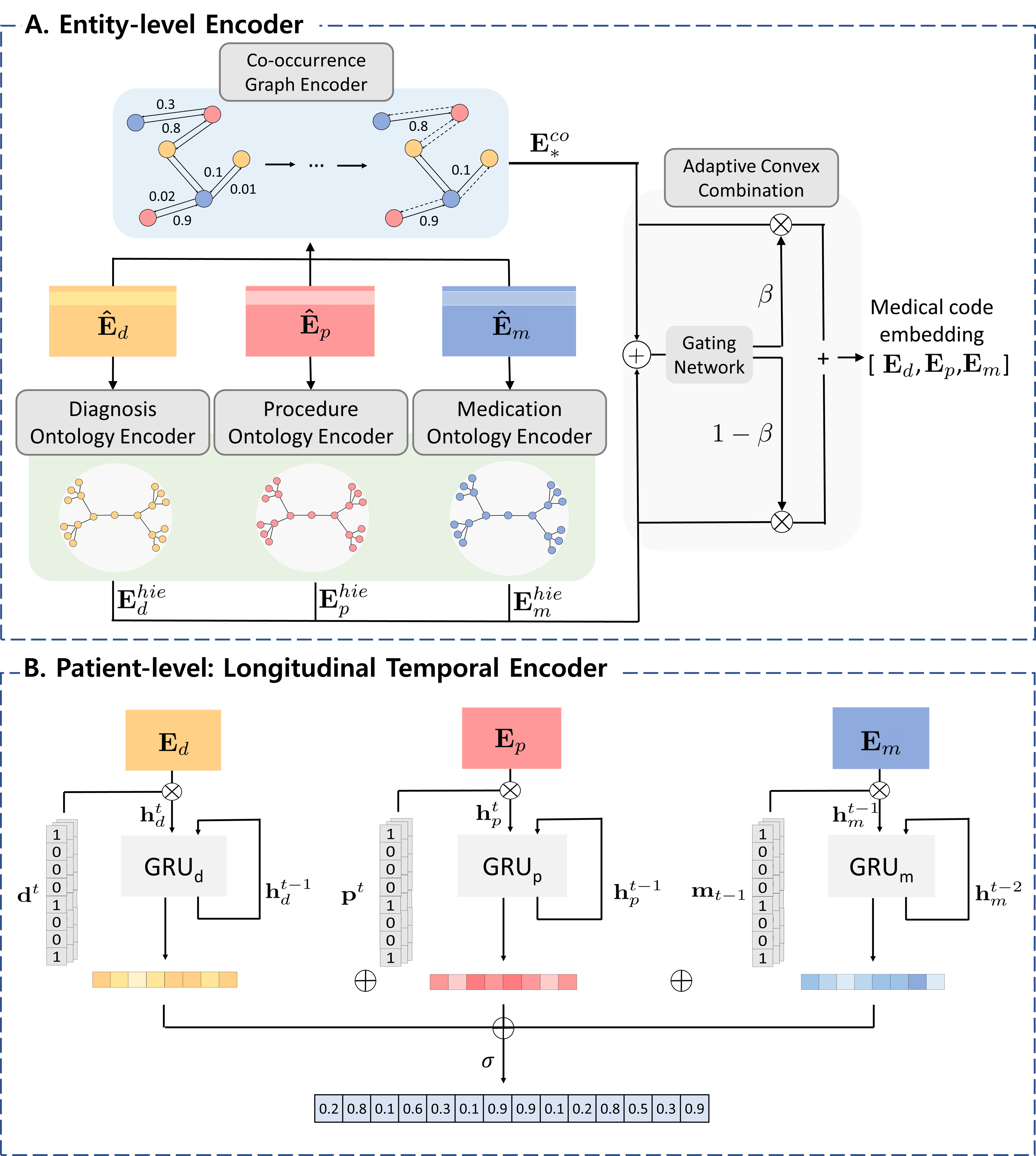}
  \caption{Overall model architecture with A) entity-level and B) patient-level representation learning modules. The former includes three submodules: 1) co-occurrence graph encoder, 2) separate hierarchical ontology encoders and 3) an adaptive convex combination gate.}
  \label{fig:modelarch}
\end{figure}

\subsection{Problem Formulation}
Let $D$,$P$, and $M$ be finite sets of diagnosis, procedure, and medication codes, respectively.
Each visit $t$ is represented by multi-label binary indicators $\mathbf{d}_t\in\{0,1\}^{|D|}$, $\mathbf{p}_t\in\{0,1\}^{|P|}$, and $\mathbf{m}_t\in\{0,1\}^{|M|}$.
For patient $i$ with visits $t=1,\ldots,T_i$, where $T_i$ is the total number of visits for patient $i$. We define the visit tuple 
$\mathbf{x}^{(i)}_t=(\mathbf{d}^{(i)}_t,\mathbf{p}^{(i)}_t,\mathbf{m}^{(i)}_t)$ 
and the history up to (but excluding) $t$ as:
\begin{equation}
    H^{(i)}_t=(\mathbf{x}^{(i)}_1,\ldots,\mathbf{x}^{(i)}_{t-1}).
\end{equation}
At visit $t$, given $(H^{(i)}_t,\mathbf{d}^{(i)}_t,\mathbf{p}^{(i)}_t)$, the goal is to estimate
\begin{equation}
\mathbb P (\mathbf m^{(i)}_t=1\ \big|\ H^{(i)}_t,\mathbf{d}^{(i)}_t,\mathbf{p}^{(i)}_t).
\end{equation}

Given a training dataset $\mathcal{S}=\{\mathbf{X}^{(i)}\}_{i=1}^N$ with $\mathbf{X}^{(i)}=(\mathbf{x}^{(i)}_1,\ldots,\mathbf{x}^{(i)}_{T_i})$, we learn $\theta$ by minimizing the loss function:
\begin{equation}
    \min_\theta\ \frac{1}{\sum_i T_i}\sum_{i=1}^N \sum_{t=1}^{T_i}
    \ell\left(\mathbf{m}^{(i)}_t,
    f_\theta(H^{(i)}_t,\mathbf{d}^{(i)}_t,\mathbf{p}^{(i)}_t)\right).
\end{equation}
where $N$ is the total number of patients in the dataset $\mathcal{S}$. For notation simplicity, we will omit the patient notation $(i)$ in the following scripts.

\section{Methods}
\ourmodel\ comprises two modules: (i) \emph{entity-level representation learning}, which learns code embeddings by encoding hierarchical ancestry and empirical co-occurrence information (Figure~\ref{fig:modelarch}A); and (ii) \emph{patient-level representation learning}, which aggregates visit sequences for medication recommendation (Figure~\ref{fig:modelarch}A).

Let $dim\in\mathbb N$ denote the shared embedding dimension.
We define three base embedding tables for the finite sets $D$, $P$, and $M$:
\begin{equation}
 \hat{\mathbf E}_d\in\mathbb R^{|D|\times dim},\quad
 \hat{\mathbf E}_p\in\mathbb R^{|P|\times dim},\quad
 \hat{\mathbf E}_m\in\mathbb R^{|M|\times dim},
\end{equation}
whose rows are the initial embeddings for each code including their ancestor code embeddings.
These embeddings are then adapted by the submodules defined below.

\subsection{Entity-Level: Hyperbolic Ontology Encoder}

Clinical ontologies (e.g., ICD, ATC) are \emph{rooted trees}, where each node has only one parent. 
We assume each ontology tree as $\mathcal T_*=(V_*,E_*)$ for each type $*\in\{d,p,m\}$ with node set $V_*$ equal to the codes of that particular type. We embed each tree into their respective hyperbolic space (i.e. Poincaré ball model)\citep{nickel2017poincare}. The distance in the space is defined as:
\begin{equation}
    {dist}_{\mathbb B}(x,y)=\operatorname{arccosh} \left(1+\frac{2\|x-y\|^2}{(1-\|x\|^2)(1-\|y\|^2)}\right).
\end{equation}
where ${dist}_{\mathbb B}(x,y)$ denotes the distance between point $x$ and $y$ in the hyperbolic space $\mathbb{B}$. To embed ontology trees into space $\mathbb{B}$, each medical code embedding is first projected by an exponential projection $\operatorname{exp}_{\mathbb{B}}:\mathbb{R}^{dim} \to \mathbb{B}^{dim}$.
To preserve ancestry, we minimize a margin-regularized objective over all directed child-ancestor pairs:
\begin{align}
    \mathcal{L}_{\text{hyp}} = \sum_{*\in\{d,p,m\}} \sum_{(i, j) \in \mathcal{P}} {dist}_{\mathbb{B}}(\mathbf{e}^{\mathbb{B}}_{*,i}, \mathbf e^{\mathbb{B}}_{*,j}), \text{where } \mathbf{e}^{\mathbb{B}}_{*,i} = \operatorname{exp}_{\mathbb{B}}((\hat{\mathbf{E}}_*)_i)
\end{align}
where $\mathcal P_*\subseteq V_*\times V_*$ contains child-ancestor pairs, while $\mathbf{e}^{\mathbb{B}}_{*,i}$ and $\mathbf{e}^{\mathbb{B}}_{*,j}$ are the hyperbolic projection of vectors indexed from $\hat{\mathbf{E}}_*$.

\begin{figure}[htbp]
  \centering
  \includegraphics[width=0.7\linewidth]{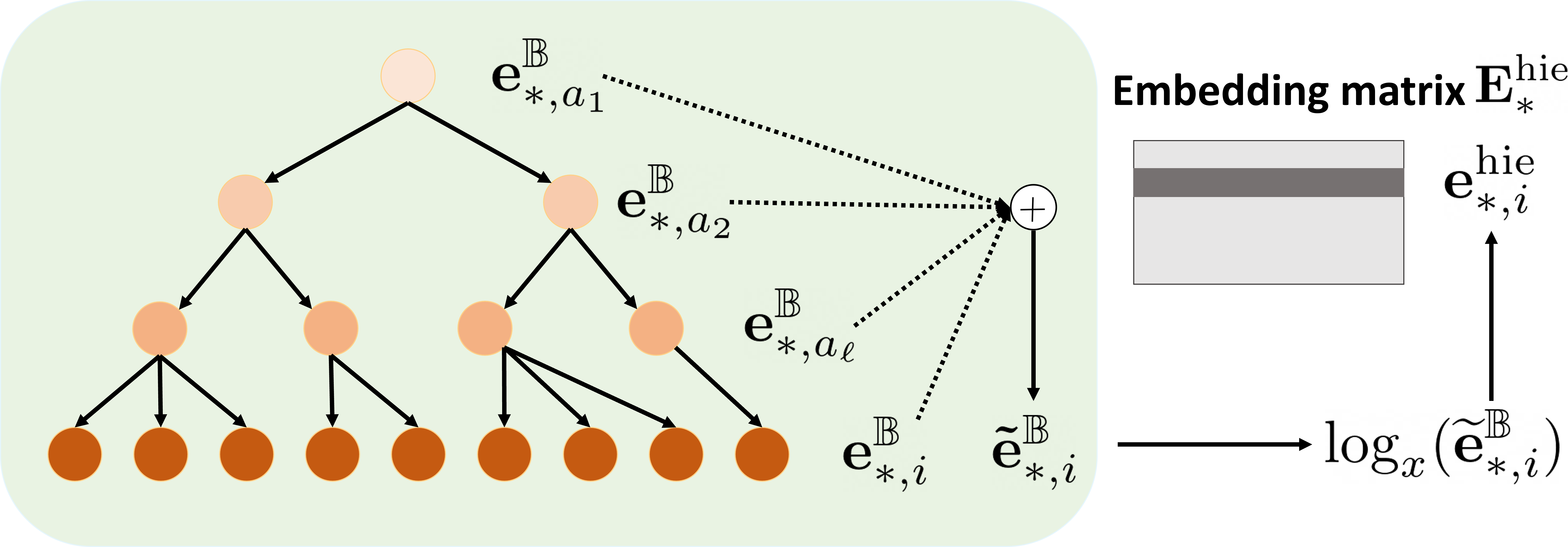}
  \caption{Hierarchical ontology encoder. }
  \label{fig:hyp}
\end{figure}

\paragraph{Möbius aggregation of ancestors.}
To more explicitly encode hierarchical information into the leaf node representation, we aggregate ancestry information into the leaf representation, inspired by the approach of GRAM~\citep{choi2017gram}. The operation is illustrated in Figure~\ref{fig:hyp}. For a node $i\in V_*$ with ordered ancestors $(a_1,\dots,a_\ell)$ from root to direct parent, we form an aggregated hyperbolic representation via Möbius addition $\oplus$, which serves as the hyperbolic analog vector addition operation in hyperbolic geometry $\mathbb B^{dim}$:
\begin{equation}
\widetilde{\mathbf{e}}^{\mathbb{B}}_{*,i}
= \mathbf{e}^{\mathbb{B}}_{*,a_1}\oplus\cdots\oplus \mathbf{e}^{\mathbb{B}}_{*,a_\ell}\oplus \mathbf{e}^{\mathbb{B}}_{*,i}.
\end{equation}
where $\mathbf{e}^{\mathbb{B}}_{*,i}$ is the medical code hyperbolic representation vector itself.

\paragraph{Euclidean compatibility via the log map.}
Subsequently, we map the hyperbolic vectors to Euclidean space using the logarithmic map at the origin $\log_x:\mathbb B^{dim}\to\mathbb R^{dim}$, where the mapped Euclidean hierarchical embeddings are defined as:
    
\begin{equation}
    \mathbf E^{\mathrm{hie}}_*=\bigl[\log_x(\widetilde{\mathbf e}^{\mathbb{B}}_{*,i})\bigr]_{i\in V_*} \in \mathbb R^{|V_*|\times d}.
\end{equation}
This ensures that the learned embeddings preserve the latent hierarchy during training while remaining compatible with Euclidean-based models for subsequent processing tasks.

\subsection{Entity-Level: Co-occurrence Graph Encoder}

While encoding hierarchical ancestry in medical code systems provides valuable semantic structure, co-occurrence patterns in EHR data offer complementary information—particularly regarding cross-entity-type interactions among diagnoses, procedures, and medications.

To fully leverage co-occurrence signals, we first construct a dense, global, directed co-occurrence graph by connecting all medical entities that co-occur within a single visit. Each pair of co-occurring entities is linked with both incoming and outgoing edges. The weight of each edge reflects the conditional probability of co-occurrence from the source node. Formally, the prior edge-weight matrix is denoted as $\mathbf{A} \in \mathbb{R}^{\mid V \mid \times \mid V \mid}$, where $\mid V \mid = |D| + |P| + |M|$ is the total number of codes. The entry of $(i,j)$ is defined as bellow:
\begin{equation}
    \mathrm{a}_{ij} = \frac{|\textit{occ}(i) \cap \textit{occ}(j)|}{|\textit{occ}(i)|} \in [0,1] ,
\end{equation}
where $\textit{occ}(i)$ denotes the set of visits in which code $i$ appears. Thus, $\mathrm{a}_{ij} \in [0,1]$ quantifies the likelihood of observing code $j$ given the presence of code $i$. 

\paragraph{Masked-softmax graph attention with sparsity.}
We adopt the sparsity framework of Sparse GAT (SGAT) \cite{ye2021sparse}—learning a stochastic binary mask $Z$ with an $L_0$-style regularizer. Note that while departing from SGAT in the normalization step: we use a \emph{masked softmax} to normalize attention over each node’s (masked) neighborhood. 
Let $h^{(0)} = [\hat{\mathbf E}_d, \hat{\mathbf E}_p, \hat{\mathbf E}_m]$ be the initial entity embeddings. 
At layer $l$, we compute raw attention scores by combining learned feature compatibility with the co-occurrence prior:
\begin{align}
    s_{ij}^{(l)} \;=\; g \big(h_i^{(l)},\,h_j^{(l)}\big) \;+\; \eta\,\log\left(p_{ij}\right), \qquad j \in \mathcal{N}_i,
\end{align}
where $g(\cdot)$ is a learnable scoring function (e.g., a GAT-style bilinear/additive scorer) that outputs a scalar value, and $\eta$ is a hyperparameter controlling the impact of prior weights on the masking probability. Given the stochastic edge mask $z_{ij}\in\{0,1\}$ (defined below), the masked-softmax attention is
\begin{align}
    \tilde{\alpha}_{ij}^{(l)} \;=\; \exp \left(\frac{s_{ij}^{(l)}}{\tau}\right), \qquad
    \alpha_{ij}^{(l)} \;=\;
    \frac{\tilde{\alpha}_{ij}^{(l)}\, z_{ij}}{\sum_{k\in\mathcal{N}_i} \tilde{\alpha}_{ik}^{(l)}\, z_{ik}} ,
\end{align}
where $\tau>0$ is a temperature controlling attention sharpness. Node updates are then defined as follows:
\begin{equation}
    h_i^{(l+1)} \;=\; \phi \left( \sum_{j\in\mathcal{N}_i} \alpha_{ij}^{(l)}\, h_j^{(l)} W^{(l)} \right),
\end{equation}
with $W^{(l)}$ a learnable weight matrix and $\phi(\cdot)$ a pointwise nonlinearity (e.g., ELU). 
Note that, unlike SGAT’s row-normalization of $A\odot Z$, our $\alpha_{ij}^{(l)}$ are \emph{softmax-normalized} over the \emph{masked} neighborhood, resulting in a convex combination of neighbors that integrates both the learned compatibility and the prior.
We denote the final co-occurrence embeddings by:
\begin{equation}
    \mathbf E^{\mathrm{co}}_*  =  \mathbf{h}^{L} \in \mathbb{R}^{|V_*|\times d} \quad 
\end{equation}

\paragraph{Objective with $L_0$ sparsity.}
Let $f_i(\cdot)$ denote the model’s prediction for sample $i$ based on the masked-softmax attention. We optimize
\begin{equation}
\hat{\mathcal{R}}(W,\kappa) \;=\; \frac{1}{n} \sum_{i=1}^{n} \mathbb{E}_{q(Z\,|\,\kappa)} \,
\mathcal{L} \big(f_i(X,\, \alpha^{(l)}(Z),\, W),\, y_i\big)
\;+\; \lambda \sum_{(u,v)\in E} \pi_{uv},
\end{equation}
where $q(Z\,|\,\kappa)$ defines a hard-concrete (relaxed Bernoulli) distribution over edge masks parameterized by $\kappa=\{\kappa_{uv}\}$, and $\pi_{uv}$ denotes the corresponding inclusion probabilities used by the $L_0$ penalty.\footnote{Equivalently, one may write the penalty in terms of $\mathbb{E}[Z_{uv}]$ under $q(Z\,|\,\kappa)$.} The expectation is taken over the randomness in $Z$; during training we use the relaxed $z_{uv}$, and at inference we can do hard-thresholding.
To encourage sparsity we penalize a proxy for the expected number of active edges: $\mathcal{L}_{\text{sparse}}=\sum_{i,j}\pi_{ij}$.

\paragraph{Hard-concrete reparameterization and prior-informed gates.}
For each potential edge $(i,j)$, we sample a relaxed gate $z_{ij}\in(0,1)$ via
\begin{align}
    u_{ij} &\sim \mathcal{U}(0,1), \\
    z_{ij} &= \sigma_g \left( \frac{1}{\beta} \left( \log u_{ij} - \log(1-u_{ij}) + \log \bar{\kappa}_{ij} \right) \right), \\
    \log \bar{\kappa}_{ij} &= \log \kappa_{ij} + \gamma\, \log(p_{ij}),
\end{align}
where $\sigma_g(\cdot)$ is the logistic sigmoid, $\beta>0$ is a temperature controlling the relaxation sharpness, $\kappa_{ij}>0$ is a learnable gate parameter, and $\gamma$ regulates how strongly the co-occurrence prior $p_{ij}$ biases edge inclusion. The masked-softmax attention uses these $z_{ij}$ to zero out pruned edges and renormalize over the remaining neighbors.

\paragraph{Why softmax-normalized attention here?}
Notably, different from the original implementation in SGAT, we apply softmax-normalized attention for the attention scores for the following reasons:
\emph{(i) Degree-robust convex combinations:} $\sum_j \alpha_{ij}^{(l)}=1$ stabilizes the scale of aggregated messages across visits with widely varying numbers of codes. 
\emph{(ii) Principled fusion of priors and features:} placing $\psi(p_{ij})$ in the logits yields a Gibbs distribution that balances data-driven compatibility with co-occurrence evidence; $\tau$ affords direct control of attention entropy.
\emph{(iii) Implicit sparsification and interpretability:} the exponential weighting amplifies confident neighbors, complementing the $L_0$ mask, and the resulting probabilities are easily interpretable in clinician-facing summaries.

\input{tables/statistics}
\subsection{Adaptive Convex Combination of Ontology- and Co-occurrence-Based Representations}
We assume that the contribution of each component from the previous submodules should vary across medical entities. The intuition is that certain entities may benefit more from co-occurrence-based representations, while others may rely more heavily on information from their ontological ancestors to capture meaningful semantics for the downstream drug recommendation task. 

Therefore, we design an adaptive convex combination layer to integrate $E^{hie}$ and $E^{co}$ to adaptively learn the contribution of each component for different medical entities. For code $i$ of type $*$,
\[
\beta_{*,i}=\mathrm{sigmoid} \big(w^\top [\,(\mathbf E^{\mathrm{hie}}_*)_{i,:}\ ;\ (\mathbf E^{\mathrm{co}}_*)_{i,:}\,]+b\big)\in(0,1),
\]
\[
(\mathbf E_*)_{i,:}
=\beta_{*,i}\,(\mathbf E^{\mathrm{hie}}_*)_{i,:}+(1-\beta_{*,i})\,(\mathbf E^{\mathrm{co}}_*)_{i,:},
\]
with shared gate parameters $w\in\mathbb R^{2d}, b\in\mathbb R$.
The resulting $\mathbf E_*\in\mathbb R^{|V_*|\times d}$ are the entity embeddings that is subsequently fed to the patient-level representation module.

\subsection{Patient-Level: Longitudinal Temporal Encoder}
For a patient with visits $t=1,\dots,T$, let the multi-hot vectors be $\mathbf d_t\in\{0,1\}^{|D|}$, $\mathbf p_t\in\{0,1\}^{|P|}$, $\mathbf m_t\in\{0,1\}^{|M|}$.
We obtain entity-specific visit embeddings by summing over active codes. For each entity type $* \in \{d, p, m\}$:
\begin{equation}
\mathbf h_*^t=\sum_{j:\ (\mathbf *_t)_j=1} (\mathbf E_*)_{j,:}\ \in\ \mathbb R^{d}
\end{equation}
Three GRUs (two layers each, hidden size $d$) encode longitudinal dynamics per entity type. To avoid label leakage at time $t$, the medication GRU is only fed up to $t - 1$:
\begin{gather}
\mathbf z_d^t=\mathrm{GRU}_d(\mathbf h_d^1,\dots,\mathbf h_d^t),\quad
\mathbf z_p^t=\mathrm{GRU}_p(\mathbf h_p^1,\dots,\mathbf h_p^t),\nonumber\\
\mathbf z_m^{t-1}=\mathrm{GRU}_m(\mathbf h_m^1,\dots,\mathbf h_m^{t-1})\ \in\ \mathbb R^{d},
\end{gather}
where each GRU returns its final hidden state for the given prefix, with $\mathbf z_m^{0}$ initialized to zeros (or a learned vector). The per-visit patient representation is defined as follows:
\[
\mathbf z_t=[\mathbf z_d^t;\ \mathbf z_p^t;\ \mathbf z_m^{t-1}]\in\mathbb R^{3d}.
\]

\subsection{Medication Recommendation Head}
We predict per-medication probabilities for each visit $t$ with a linear layer followed by the sigmoid:
\begin{gather}
    \hat{\mathbf m}_t=\mathrm{sigmoid}(W\mathbf{z}_t+\mathbf{b})\in[0,1]^{|M|},\\
    W\in\mathbb{R}^{|M|\times 3d},\ \mathbf{b}\in\mathbb{R}^{|M|}.
\end{gather}

\subsection{Learning Objective: Multi-Label and Structural Losses}
Let the dataset be $\mathcal S=\{\mathbf X^{(i)}\}_{i=1}^N$ with patient $i$ having $T_i$ visits.
At visit $t$ we denote ground-truth label as $\mathbf m^{(i)}_t\in\{0,1\}^{|M|}$ and prediction as $\hat{\mathbf m}^{(i)}_t\in[0,1]^{|M|}$.  We optimize all learnable parameter using the main loss functions binary cross-entropy (BCE) loss $\mathcal{L}_{bce}$ and multi-label margin loss$\mathcal{L}_{multi}$, which are defined as: 
\begin{gather}
    \mathcal{L}_{\text{bce}} = -\sum_{i=1}^{|M|} m_i \log(\hat{m}_i) + (1 - m_i) \log(1 - \hat{m}_i) \\
    \mathcal{L}_{\text{margin}} = \sum_{i,j; m_i = 1, m_j = 0} \frac{\max(0,\ 1 - (\hat{m}_i - \hat{m}_j))}{|M|}
\end{gather}

On top of that, We reuse $\mathcal L_{\text{hyp}}$ and $\mathcal L_{\text{sparse}}$ from the submodules above as the regularization terms to enforce tree structure in hyperbolic space and to induce sparsity in the co-occurrence graph, respectively.
The total objective is given by follows:
\begin{equation}
    \mathcal L=\lambda_{\text{bce}}\,\mathcal{L}_{\text{BCE}}+\lambda_{\text{margin}}\,\mathcal L_{\text{margin}}
    +\lambda_{\text{hyp}}\,\mathcal L_{\text{hyp}}+\lambda_{\text{sparse}}\,\mathcal L_{\text{sparse}},
\end{equation}
with nonnegative hyperparameters $\lambda_{\text{bce}},\lambda_{\text{margin}},\lambda_{\text{hyp}},\lambda_{\text{sparse}}\ge 0$.

\subsection{Implementation Details}
\ourmodel~was implemented in PyTorch and trained on a single NVIDIA GeForce RTX 3090 GPU with 24GB memory.
We train our model with supervision on the overall loss $\mathcal{L}$ with a maximum epochs of 200 and patience of 30, with learning rate of $1e^{-2}$. The average number of epochs before convergence is 70. We run the model with learning rate of 1e-2. 
The loss hyperparameters $\lambda_{\text{bce}}$, $\lambda_{\text{margin}}$, $\lambda_{\text{hyp}}$ and $\lambda_{\text{sparse}}$ are 0.99, 0.04, 0.01, and 0.01, respectively.

\section{Experiments}

\input{tables/table1.tex}
\subsection{Dataset, Baselines and Evaluation Metrics}
\subsubsection{Datasets} 
We utilize electronic health records (EHRs) from \textbf{MIMIC-III}\citep{johnson2016mimic} and \textbf{MIMIC-IV}\citep{johnson2023mimic}, two publicly available critical care databases widely used in clinical machine learning research. 
To ensure fair comparison, we follow the preprocessing steps used in SafeDrug for the general setting, filtering out single-visit patients and drugs without SMILES mappings. It is important to note that MIMIC-IV contains both ICD-9 and ICD-10 diagnosis and procedure codes ~\citep{hirsch2016icd}. However, due to the lack of reliable one-to-one mappings between ICD-9 and ICD-10, we retain only the ICD-9 coded diagnoses and procedures in MIMIC-IV to maintain consistency and avoid ambiguity. The statistics of MIMIC-III and MIMIC-IV are summarized in Table~\ref{tab:stats}.

\subsubsection{Baselines}

We compare our method against a range of baselines, including traditional classifiers, sequential models, and graph-based methods that incorporate structural or external knowledge.

\begin{itemize}
    \item \textbf{Logistic Regression (LR)}~\citep{luaces2012binary}: A multilabel logistic regression model using binary relevance and L2 regularization.

    \item \textbf{ECC}~\citep{read2009classifier}: An ensemble of classifier chains for multilabel prediction with label dependency modeling.

    \item \textbf{RETAIN}~\citep{choi2016retain}: A sequential model with reverse-time attention for interpretable medication prediction.

    \item \textbf{LEAP}~\citep{zhang2017leap}: A generative model using a recurrent decoder with attention and reinforcement learning.

    \item \textbf{GAMENet}~\citep{shang2019gamenet}: A memory-based model combining patient history with a DDI knowledge graph via GCN.

    \item \textbf{SafeDrug}~\citep{yang2021safedrug}: A graph-based model that encodes molecular structures and controls DDI via a custom loss.

    \item \textbf{MICRON}~\citep{yang2021change}: A residual recurrent model for medication change prediction using incremental updates.

    \item \textbf{MoleRec}~\citep{yang2023molerec}: A substructure-aware model that links drug parts to diseases and regulates DDI via annealed training.

    \item \textbf{Carmen}~\citep{chen2023context}: A graph-based model that integrates patient history and explicitly encodes DDI information.

    \item \textbf{LAMRec}~\citep{tang2024lamrec}: A label-aware multi-view model using diagnosis and procedure via cross-attention and drug label knowledge through label-wise attention and multi-view contrastive learning.
\end{itemize}

\subsubsection{Evaluation Metrics}
We evaluate model performance using the following standard metrics for multilabel medication prediction:

\begin{itemize}
    \item \textbf{Jaccard Similarity}: Measures the overlap between the predicted and ground-truth medication sets:
    \[
    \text{Jaccard} = \frac{|\mathbf{1}_{\{\hat{m}_t \geq \tau\}} \cap m_t|}{|\mathbf{1}_{\{\hat{m}_t \geq \tau\}} \cup m_t|}.
    \]
    where $\hat{m}_t$ is the predicted probability vector, $m_t$ is the ground-truth binary vector, and $\tau=0.5$ is the binarization threshold.    

    \item \textbf{PRAUC (Precision–Recall AUC)}: This score is the area under the precision–recall curve computed from the predicted medication scores and is especially useful in the presence of label imbalance.
    
    \item \textbf{F1 Score}: The harmonic mean of precision and recall computed in a micro-averaged fashion over all predicted medications across visits. 
\end{itemize}

\subsection{In-Distribution Performance Comparisons}
We evaluated the performance of \ourmodel~under general settings. 
As shown in Table~\ref{tab:main}, \ourmodel~ surpasses all baseline models on MIMIC-IV with a Jaccard score of 0.4989, PRAUC of 0.7592, and F1 score of 0.6500, while achieving superior performance compared to the baseline models on MIMIC-III. This demonstrates its overall effectiveness in recommending medications that align with real-world clinical practices. To assess the contribution of each component, we conducted ablation studies with three variants: i) \emph{\ourmodel~w/o hie} removes the hierarchical ontology encoder, ii) \emph{\ourmodel~w/o co} excludes the co-occurrence graph encoder, and iii) \emph{\ourmodel~w/o fus} replaces the convex combination gate with a simple average of the component outputs. The ablation results indicate that all modules are essential for overall performance. Notably, removing the co-occurrence graph encoder (\emph{\ourmodel~w/o co}) exhibits the most significant performance drop, underscoring the importance of capturing non-spurious, data-driven associations critical for accurate medication recommendation. The performance decline observed in removing hierarchical ontology encoder (\emph{\ourmodel~w/o hie}) highlights the value of enriching entity representations with hierarchical semantic information, thereby improving robustness and generalization. Meanwhile, although removing adaptive gate (\emph{\ourmodel~w/o fus}) results in only a small performance drop, it still demonstrates the benefit of adaptive fusion of the outputs from different modules rather than treating them equally.


\subsection{Evaluation on Unseen Settings}

\input{tables/unseen}

\begin{figure*}[htbp]
  \centering
  \includegraphics[width=0.95\linewidth]{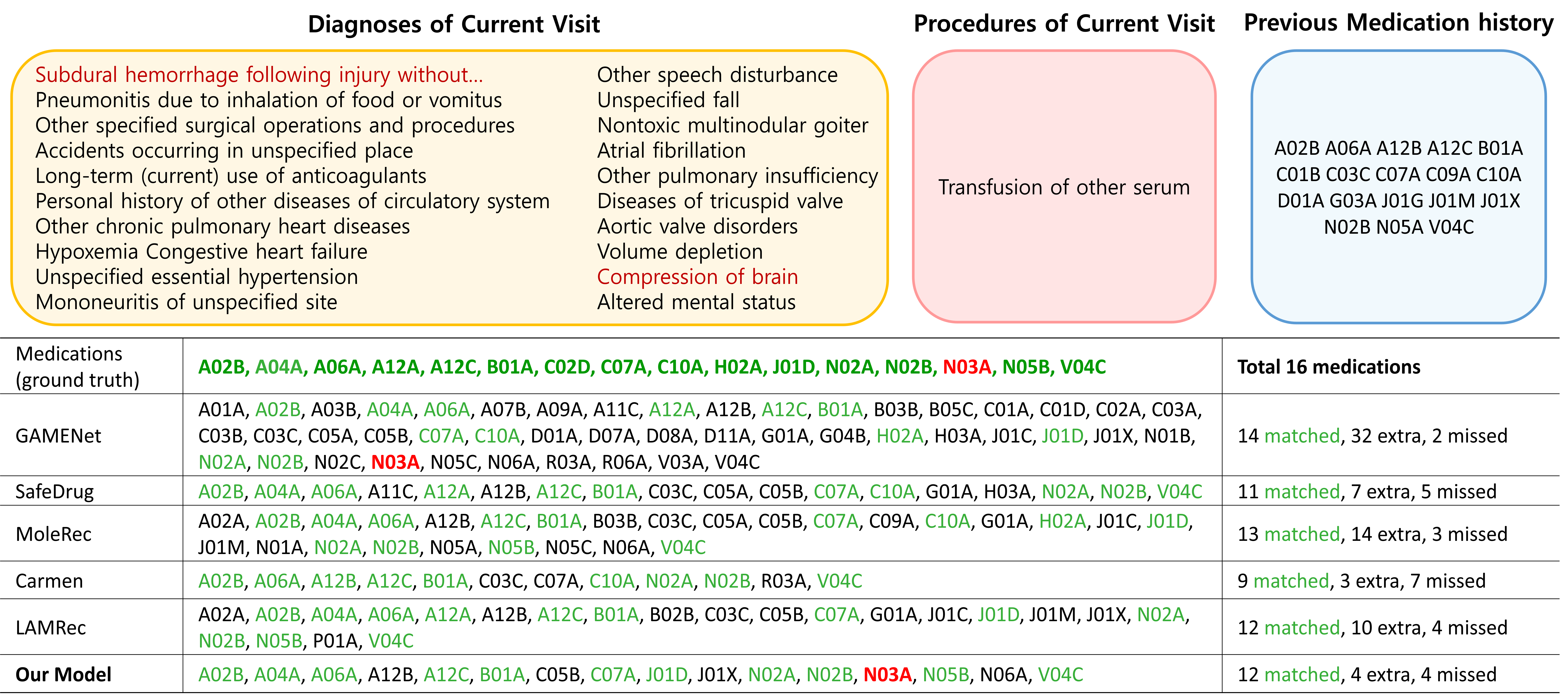}
  
  \caption{Case Study of Unseen Settings}
  \label{fig:casestudy}
\end{figure*}
We design an experiment under a challenging \emph{unseen setting}. We remove a subset of medical entities strongly correlated with a target medication from the training data, then see if the model can still recommend the right medications. The criteria of selecting strongly correlated target entities are:
1) co-occurrence rate from source to target > 0.5, 2) co-occurrence rate from target to source above 0.01. The \emph{unseen setting} tests whether the model can infer meaningful representations without relying on co-occurrence signals in the dataset. This simulates real clinical situations like encountering new rare diseases or dealing with incomplete patient records. 
We test three specific medications: \texttt{R03A} (bronchodilators for respiratory diseases), \texttt{N03A} (antiepileptics for neurological conditions), and \texttt{C01C} (cardiac stimulants for heart conditions). These medications were chosen because each is clearly associated with distinct medical domains. We compare \ourmodel's performance against several competitive baselines.

As shown in Table~\ref{tab:unseen}, \ourmodel~ consistently achieves the highest tF1 scores for all target medications. This shows that \ourmodel~can generalize using ontological structure rather than relying on co-occurrence patterns. In contrast, baseline models such as Carmen and MoleRec struggle with unseen entities because they depend heavily on correlations seen during training. 

Figure~\ref{fig:casestudy} shows a representative case where we compare the medication predictions across different models. We designate \textit{'Compression of brain'} as unseen diagnosis and mask related entities like \textit{'Subdural hemorrhage following injury'} to eliminate co-occurrence signals. Under this constraint, the model must rely on a robust embedding of \textit{'Compression of brain'} to recommend \texttt{N03A}. Most baseline models fail this challenge, but \ourmodel~ successfully identifies \texttt{NO3A}. GAMENET, though correctly identifies the target medication, the low precision of the model indicates poor confidence and numerous false positives. This case highlights \ourmodel's ability to generalize using hierarchical ontological relationships as complement to the occurrence patterns, especially under unseen settings and underscores its potential for reliable medication recommendation in sparse or unseen clinical scenarios.
\begin{figure*}[htbp]
  \centering
  \includegraphics[width=1\linewidth]{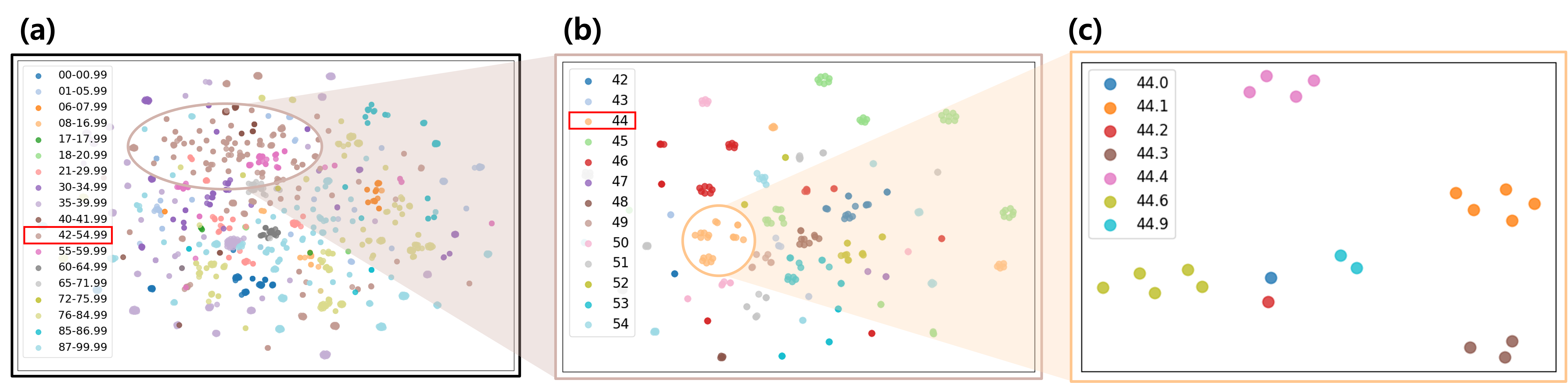}
  \caption{Learned Embeddings visualized on TSNE plots.}
  \label{fig:embed}
\end{figure*}

\begin{figure}[htbp]
  \centering

  \includegraphics[width=0.65\linewidth]{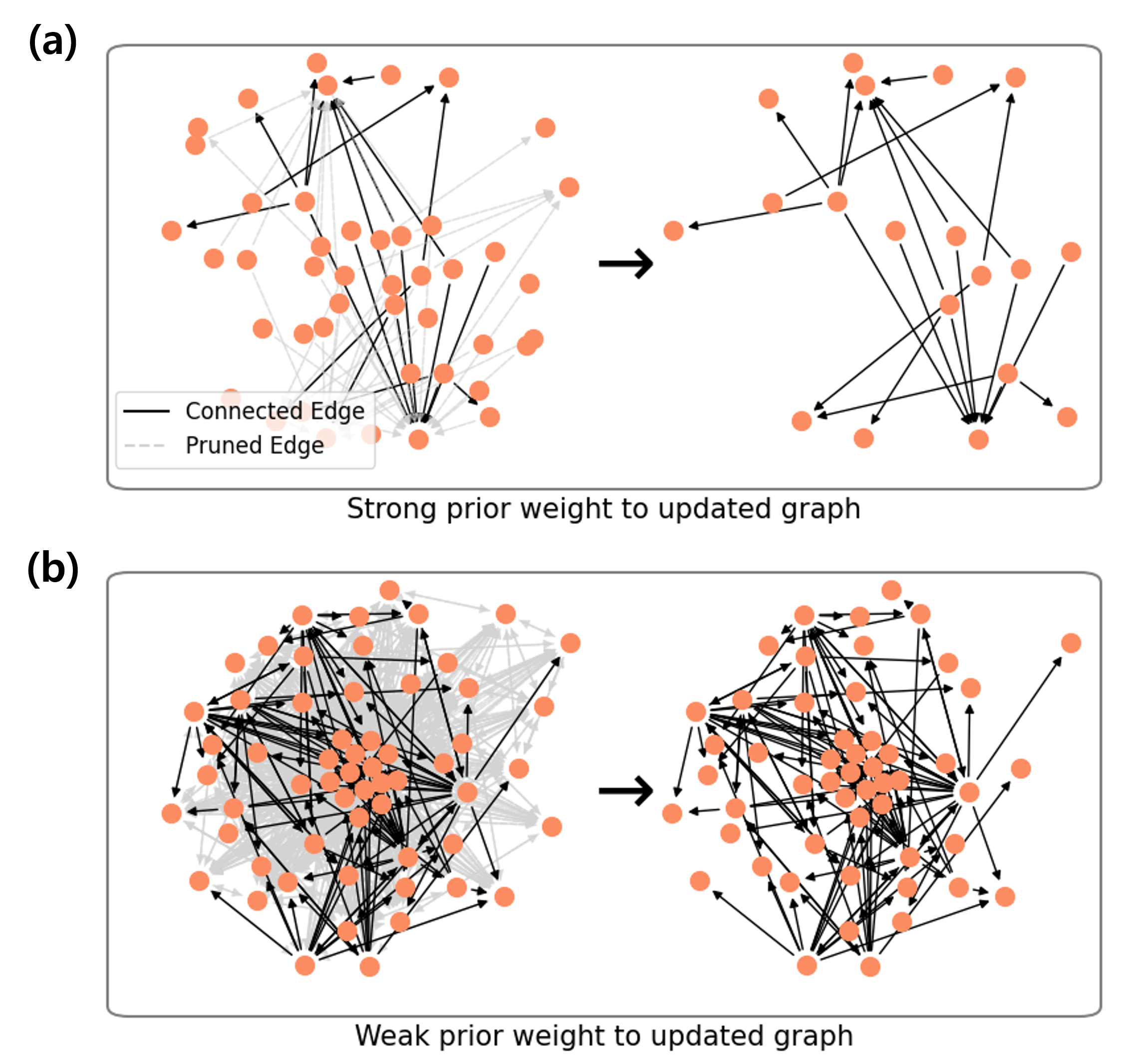}
  \caption{Sparse graph analysis}
  \label{fig:graph}
\end{figure}

\subsection{Learning Edge Refinement}
We refined the graph structure to be sparser, aiming to suppress spurious correlations, while improving both computational efficiency and interpretability. To assess whether the model updates the graph in an explainable manner, we analyzed three representative scenarios.

First, we examined whether edges with high prior edge weights were retained after training (Fig.~\ref{fig:graph}a). 
We observed that such edges initially deemed informative for drug recommendation tend to be preserved during learning. 
For example, in patients receiving gastrointestinal medications, the model maintains connections between  \textit{'parasympathomimetics'} and \textit{'drugs for peptic ulcer and GORD'}, and between \textit{'drugs for peptic ulcer and GORD'} and \textit{'drugs for functional gastrointestinal disorders'}. This indicates that the graph is both clinically meaningful and interpretable.

Second, we investigated cases where edges with low prior edge weights are preserved (Fig.~\ref{fig:graph}b). Despite weak initial connections, the model identifies these as useful for drug recommendation. For instance, edges between \textit{'antithrombotic agents'} and conditions like \textit{'hydronephrosis'} or \textit{'antiglaucoma preparations and miotics'} have low co-occurrence rates. However, the model retains these edges and links them to meaningful drug recommendations such as \textit{'potassium-sparing diuretics'} and \textit{'hypothalamic hormones'}. This suggests the model discovers clinically relevant associations beyond simple co-occurrence patterns.

Finally, we analyzed edges that are pruned during training despite having high prior weights. Although these edges show strong co-occurrence patterns, they contribute little to accurate prediction. The pruned target nodes commonly exhibit high out-degree values, indicating broad but non-specific connectivity. For example, nodes such as \textit{'drugs for constipation'} or \textit{'potassium supplements'} frequently appear with many other medical entities but provide limited discriminative power for drug recommendation. To quantitatively support this observation, we compared the out-degree distributions between all nodes and target nodes of pruned edges. Target nodes of pruned edges had a significantly higher median out-degree (529 vs. 268; Mann–Whitney U test, $p < 0.0001$). 
This suggests that highly connected nodes may be less informative for learning specific treatment associations, despite being prevalent.

Overall, these analyses demonstrate that the model learns to retain clinically meaningful edges while discarding uninformative ones. This selective approach improves both performance and interpretability, key requirements for clinical decision support.

\subsection{Visualization of Learned Embeddings}
We examine the properties of learned embeddings through t-SNE visualization in Figure~\ref{fig:embed}. Figure~\ref{fig:embed} present embeddings colored by hierarhical ontology levels of procedure codes. We focus on procedure codes because they have minimal co-occurrence edges among themselves due to their sparse occurrence in patient visits. As a result, their embeddings are more strongly influenced by the underlying medical ontology structure. The average value of the convex fusion gate $\beta$ for procedure codes is 0.89, indicating a strong contribution from ontology-based learning. 
The visualizations reveal that embeddings from the same ontological subgroup generally cluster together across different hierarchy levels. However, some inter-group overlap and dispersion occur, reflecting the influence of co-occurrence patterns learned from patient data. These findings highlight \ourmodel's effectiveness in capturing ontological relationships while balancing data-driven co-occurrence patterns.

\section{Conclusion}
In this paper, we introduced \ourmodel, a robust medication recommendation framework that leverages two complementary structural sources: hierarchical medical ontologies and data-driven EHR co-occurrence graphs. Through hyperbolic embedding of clinical codes and knowledge transfer via shared ancestors, \ourmodel~ achieves generalization to unseen medical codes. The global co-occurrence graph further captures clinically meaningful associations while filtering spurious correlations. Comprehensive experiments on MIMIC-III and MIMIC-IV validate that \ourmodel~ delivers strong in-distribution performance and maintains accuracy under distributional shifts and clinical constraints. However, we acknowledge several limitations of the current approach. First, the current framework excludes patient attributes like age and gender, which are important factors in medication selection. Second, it excludes auxiliary input modalities such as lab results and clinical notes. Third the framework lacks explicit temporal modeling of visit sequences and intervals. Finally, coding system inconsistencies between ICD-9 and ICD-10 restrict our evaluation to ICD-9 representations only.  Future work will address these limitations by incorporating patient demographics and multimodal clinical data, developing temporal sequence modeling capabilities, and enabling cross-ontology compatibility.


\bibliographystyle{unsrtnat}
\bibliography{main}  





\clearpage
\appendix
\section{Appendix}

\subsection{Data Preprocessing}
Following the widely used preprocessing steps from~\citep{yang2021safedrug},we used the publicly available MIMIC-III and MIMIC-IV dataset. Diagnosis, procedure, and medication data were extracted from the original files and merged by patient and visit IDs. ICD-9 codes were transformed into multi-hot vectors. For DDI information, we used the Top-40 severity types from TWOSIDES (ATC 3rd level), converting NDC codes to ATC and aligning drug-level molecular features accordingly using DrugBank and RDKit. We further filtered out patients with only one visit and retaining those with two or more. For MIMIC-IV, we used only ICD-9 coded data under the same processing pipeline.

\subsection{Drug-Drug Interaction}
Previous studies report the number of adverse drug-drug interactions (DDIs) as defined by the TWOSIDES database~\citep{tatonetti2012data} as DDI rate. However, as mentioned in~\citep{chen2023context} we note that this score may not reliably reflect real-world safety in the MIMIC dataset. This is because TWOSIDES defines side effects based on statistical associations, and many drug combinations labeled as adverse in TWOSIDES do, in fact, appear in the real-world prescriptions recorded in MIMIC-III/IV. Consequently, a model that strictly avoids such combinations (e.g., through DDI loss or encoding) may compromise performance by contradicting patterns observed in real-world clinical practice. Due to this misalignment, the DDI rate may not be a meaningful or comprehensive metric for evaluating safety in this setting and was excluded for our main table in the paper. We additionally reported aDDI from~\citep{chen2023context}, which measures the number of adverse DDIs not present in the prescriptions of the test dataset. Also, wDDI and awDDI also involves the confidence score from TWOSIDES database to give higher weights to drug pairs that have high confidence scores.

\subsection{Poincaré Ball Model}

In the hierarchical ontology encoder, we map Euclidean embeddings of ontology concepts into the Poincaré ball model of hyperbolic space to better capture hierarchical relationships. This projection from Euclidean to hyperbolic space is performed using the exponential map, defined as follows:

\begin{equation}
\operatorname{exp}_{\mathbb{B}}:\mathbb{R}^{dim} \to \mathbb{B}^{dim}, \quad
\operatorname{exp}_{\mathbb{B}}(x) =
\begin{cases}
\displaystyle \frac{x}{\|x\| - \varepsilon} \, & \text{if}~  \|x\| \ge 1, \\[6pt]
x & \text{otherwise},
\end{cases}
\end{equation}

where $x$ is a vector in Euclidean space, and $\varepsilon$ is a small positive constant to prevent numerical instability near the boundary of the Poincaré ball. This ensures that the projected embeddings lie within the unit ball, preserving the geometry of hyperbolic space.

To map vectors from the Poincaré ball back to Euclidean space, we use the logarithmic map, which serves as the inverse operation:

\begin{equation}
\log_x:\mathbb{B}^{dim} \to \mathbb{R}^{dim}, \quad
\log_x(y) =
2\operatorname{arctanh}(|{-x \oplus y}|) \cdot \frac{{-x \oplus y}}{|{-x \oplus y}|},
\end{equation}

where $\oplus$ denotes the Möbius addition defined in hyperbolic geometry. The logarithmic map provides a way to compute tangent vectors in the Euclidean space corresponding to points in the hyperbolic manifold, enabling gradient-based optimization in Riemannian space.

\clearpage

\subsection{Additional Results}
\input{tables/appendix_ddi_mimic3.tex}
\input{tables/appendix_ddi_mimic4.tex}

\clearpage

\subsection{Overall Performance Comparison for Unseen Settings}
\input{tables/appendix_unseen_R03A.tex}

\clearpage

\subsection{Graph Analysis}

\input{tables/appendix_graph_analysis.tex}

\end{document}

%% file: tables/statistics.tex
\begin{table}[]

\centering
\label{tab:stats}
{
\begin{tabular}{lrr}
\hline
                        & \multicolumn{1}{l}{\textbf{MIMIC III}} & \multicolumn{1}{l}{\textbf{MIMIC IV}} \\ \hline
\#patient               & 5443                          & 49885                        \\
\#visit                 & 14126                         & 115971                       \\
\#visits per patient    & 2.59                          & 2.32                         \\
\#unique diagnosis      & 1956                          & 941                          \\
\#unique procedure      & 1408                          & 668                          \\
\#unique medication     & 131                           & 131                          \\
\#diagnosis per sample  & 30.7                          & 32.9                         \\
\#procedure per sample  & 9.4                           & 7.03                         \\
\#medication per sample & 24.2                          & 18.12                        \\ \hline
\end{tabular}
}
\caption{Statistics of datasets after preprocessing.}

\end{table}

%% file: tables/table1.tex
\begin{table*}[]
\label{tab:main}
\resizebox{\textwidth}{!}
{
\small
\begin{tabular}{c|ccc|ccc}
\hline
                   & \multicolumn{3}{c|}{\textbf{MIMIC III}}                                                                                                                                                                      & \multicolumn{3}{c}{\textbf{MIMIC IV}}                                                                                                                                                                        \\ \cline{2-7} 
                   & \textbf{Jaccard $\uparrow$}                                                   & \textbf{PRAUC $\uparrow$}                                                     & \textbf{F1 $\uparrow$}                                                        & \textbf{Jaccard $\uparrow$}                                                   & \textbf{PRAUC $\uparrow$}                                                     & \textbf{F1 $\uparrow$}                                                        \\ \hline
LR                 & \begin{tabular}[c]{@{}c@{}}0.4594  $\pm$ 0.0058\end{tabular}          & \textbf{\begin{tabular}[c]{@{}c@{}}0.7413  $\pm$ 0.0053\end{tabular}} & \begin{tabular}[c]{@{}c@{}}0.6172  $\pm$ 0.0057\end{tabular}          & \begin{tabular}[c]{@{}c@{}}0.4633  $\pm$ 0.0028\end{tabular}          & \begin{tabular}[c]{@{}c@{}}0.7560  $\pm$ 0.0024\end{tabular}          & \begin{tabular}[c]{@{}c@{}}0.6135  $\pm$ 0.0025\end{tabular}          \\
ECC                & \begin{tabular}[c]{@{}c@{}}0.4466  $\pm$ 0.0056\end{tabular}          & \begin{tabular}[c]{@{}c@{}}0.7389  $\pm$ 0.0056\end{tabular}          & \begin{tabular}[c]{@{}c@{}}0.6012  $\pm$ 0.0058\end{tabular}          & \begin{tabular}[c]{@{}c@{}}0.4366  $\pm$ 0.0030\end{tabular}          & \begin{tabular}[c]{@{}c@{}}0.7465  $\pm$ 0.0024\end{tabular}          & \begin{tabular}[c]{@{}c@{}}0.5829  $\pm$ 0.0027\end{tabular}          \\
Retain             & \begin{tabular}[c]{@{}c@{}}0.4342  $\pm$ 0.0049\end{tabular}          & \begin{tabular}[c]{@{}c@{}}0.7153  $\pm$ 0.0061\end{tabular}          & \begin{tabular}[c]{@{}c@{}}0.5959  $\pm$ 0.0051\end{tabular}          & \begin{tabular}[c]{@{}c@{}}0.4315  $\pm$ 0.0032\end{tabular}          & \begin{tabular}[c]{@{}c@{}}0.7171  $\pm$ 0.0032\end{tabular}          & \begin{tabular}[c]{@{}c@{}}0.5873  $\pm$ 0.0032\end{tabular}          \\
Leap               & \begin{tabular}[c]{@{}c@{}}0.3806  $\pm$ 0.0061\end{tabular}          & \begin{tabular}[c]{@{}c@{}}0.5523  $\pm$ 0.0079\end{tabular}          & \begin{tabular}[c]{@{}c@{}}0.5399  $\pm$ 0.0064\end{tabular}          & \begin{tabular}[c]{@{}c@{}}0.3744  $\pm$ 0.0100\end{tabular}          & \begin{tabular}[c]{@{}c@{}}0.5007  $\pm$ 0.0141\end{tabular}          & \begin{tabular}[c]{@{}c@{}}0.5257  $\pm$ 0.0104\end{tabular}          \\
GAMENet            & \begin{tabular}[c]{@{}c@{}}0.4710  $\pm$ 0.0049\end{tabular}          & \begin{tabular}[c]{@{}c@{}}0.7269  $\pm$ 0.0056\end{tabular}          & \begin{tabular}[c]{@{}c@{}}0.6275  $\pm$ 0.0043\end{tabular}          & \begin{tabular}[c]{@{}c@{}}0.4795  $\pm$ 0.0048\end{tabular}          & \begin{tabular}[c]{@{}c@{}}0.7314  $\pm$ 0.0045\end{tabular}          & \begin{tabular}[c]{@{}c@{}}0.6350  $\pm$ 0.0054\end{tabular}          \\
SafeDrug           & \begin{tabular}[c]{@{}c@{}}0.4643  $\pm$ 0.0064\end{tabular}          & \begin{tabular}[c]{@{}c@{}}0.7352  $\pm$ 0.0076\end{tabular}          & \begin{tabular}[c]{@{}c@{}}0.6230  $\pm$ 0.0065\end{tabular}          & \begin{tabular}[c]{@{}c@{}}0.4351  $\pm$ 0.0095\end{tabular}          & \begin{tabular}[c]{@{}c@{}}0.6971  $\pm$ 0.0096\end{tabular}          & \begin{tabular}[c]{@{}c@{}}0.5878  $\pm$ 0.0092\end{tabular}          \\
MICRON             & \begin{tabular}[c]{@{}c@{}}0.4736  $\pm$ 0.0028\end{tabular}          & \begin{tabular}[c]{@{}c@{}}0.7274  $\pm$ 0.0094\end{tabular}          & \begin{tabular}[c]{@{}c@{}}0.6311  $\pm$ 0.0026\end{tabular}          & \begin{tabular}[c]{@{}c@{}}0.4643  $\pm$ 0.0038\end{tabular}          & \begin{tabular}[c]{@{}c@{}}0.7085  $\pm$ 0.0036\end{tabular}          & \begin{tabular}[c]{@{}c@{}}0.6155  $\pm$ 0.0034\end{tabular}          \\
MoleRec            & \begin{tabular}[c]{@{}c@{}}0.4802  $\pm$ 0.0081\end{tabular}          & \begin{tabular}[c]{@{}c@{}}0.7362  $\pm$ 0.0174\end{tabular}          & \begin{tabular}[c]{@{}c@{}}0.6374  $\pm$ 0.0077\end{tabular}          & \begin{tabular}[c]{@{}c@{}}0.4664  $\pm$ 0.0030\end{tabular}          & \begin{tabular}[c]{@{}c@{}}0.7254  $\pm$ 0.0033\end{tabular}          & \begin{tabular}[c]{@{}c@{}}0.6185  $\pm$ 0.0027\end{tabular}          \\
Carmen             & \begin{tabular}[c]{@{}c@{}}0.4616  $\pm$ 0.0280\end{tabular}          & \begin{tabular}[c]{@{}c@{}}0.7111  $\pm$ 0.0292\end{tabular}          & \begin{tabular}[c]{@{}c@{}}0.6196  $\pm$ 0.0263\end{tabular}          & \begin{tabular}[c]{@{}c@{}}0.4847  $\pm$ 0.0032\end{tabular}          & \begin{tabular}[c]{@{}c@{}}0.7441  $\pm$ 0.0036\end{tabular}          & \begin{tabular}[c]{@{}c@{}}0.6374  $\pm$ 0.0029\end{tabular}          \\
LAMRec             & \begin{tabular}[c]{@{}c@{}}0.4700  $\pm$ 0.0054\end{tabular}          & \begin{tabular}[c]{@{}c@{}}0.7289  $\pm$ 0.0106\end{tabular}          & \begin{tabular}[c]{@{}c@{}}0.6273  $\pm$ 0.0051\end{tabular}          & \begin{tabular}[c]{@{}c@{}}0.4790  $\pm$ 0.0022\end{tabular}          & \begin{tabular}[c]{@{}c@{}}0.7435  $\pm$ 0.0018\end{tabular}          & \begin{tabular}[c]{@{}c@{}}0.6311  $\pm$ 0.0017\end{tabular}          \\ \hline
\ourmodel \footnotesize{w/o hie} & \begin{tabular}[c]{@{}c@{}}0.4720  $\pm$ 0.0107\end{tabular}          & \begin{tabular}[c]{@{}c@{}}0.7354  $\pm$ 0.0121\end{tabular}          & \begin{tabular}[c]{@{}c@{}}0.6294  $\pm$ 0.0100\end{tabular}          & \begin{tabular}[c]{@{}c@{}}0.4792  $\pm$ 0.0107\end{tabular}          & \begin{tabular}[c]{@{}c@{}}0.7375  $\pm$ 0.0121\end{tabular}          & \begin{tabular}[c]{@{}c@{}}0.6315  $\pm$ 0.0100\end{tabular}          \\
\ourmodel \footnotesize{w/o co} & \begin{tabular}[c]{@{}c@{}}0.4609  $\pm$ 0.0101\end{tabular}          & \begin{tabular}[c]{@{}c@{}}0.7263  $\pm$ 0.0103\end{tabular}          & \begin{tabular}[c]{@{}c@{}}0.6187  $\pm$ 0.0103\end{tabular}          & \begin{tabular}[c]{@{}c@{}}0.4684  $\pm$ 0.0101\end{tabular}          & \begin{tabular}[c]{@{}c@{}}0.7287  $\pm$ 0.0103\end{tabular}          & \begin{tabular}[c]{@{}c@{}}0.6216  $\pm$ 0.0103\end{tabular}          \\
\ourmodel \footnotesize{w/o fus}   & \begin{tabular}[c]{@{}c@{}}0.4827  $\pm$ 0.0066\end{tabular}          & \begin{tabular}[c]{@{}c@{}}0.7361  $\pm$ 0.0069\end{tabular}          & \textbf{\begin{tabular}[c]{@{}c@{}}0.6391  $\pm$ 0.0065\end{tabular}}          & \begin{tabular}[c]{@{}c@{}}0.4946  $\pm$ 0.0066\end{tabular}          & \begin{tabular}[c]{@{}c@{}}0.7562  $\pm$ 0.0069\end{tabular}          & \begin{tabular}[c]{@{}c@{}}0.6459  $\pm$ 0.0065\end{tabular}          \\ \hline
\ourmodel               & \textbf{\begin{tabular}[c]{@{}c@{}}0.4832  $\pm$ 0.0066\end{tabular}} & \begin{tabular}[c]{@{}c@{}}0.7378  $\pm$ 0.0043\end{tabular}          & \begin{tabular}[c]{@{}c@{}}0.6390  $\pm$ 0.0058\end{tabular} & \textbf{\begin{tabular}[c]{@{}c@{}}0.4989  $\pm$ 0.0066\end{tabular}} & \textbf{\begin{tabular}[c]{@{}c@{}}0.7592  $\pm$ 0.0043\end{tabular}} & \textbf{\begin{tabular}[c]{@{}c@{}}0.6500  $\pm$ 0.0058\end{tabular}} \\ \hline
\end{tabular}
}
\caption{Performance comparison evaluated on MIMIC-III and MIMIC-IV.}
\end{table*}

%% file: tables/unseen.tex
\begin{table}[]
\label{tab:unseen}
\centering
\resizebox{0.6\columnwidth}{!}{
\begin{tabular}{c|ccc}
\hline
                   & \textbf{R03A}            & \textbf{N03A}            & \textbf{C01C}            \\ \hline
\textbf{GAMENet}   & 0.6283 $\pm$ 0.0575          & 0.5279 $\pm$ 0.0308          & 0.6225 $\pm$ 0.0428                         \\
\textbf{SafeDrug}  & 0.5766 $\pm$ 0.0577          & 0.4440 $\pm$ 0.0425          & 0.4709 $\pm$ 0.0680          \\
\textbf{MoleRec}   & 0.6635 $\pm$ 0.0147          & 0.5109 $\pm$ 0.0790          & 0.5094 $\pm$ 0.0598          \\
\textbf{Carmen}    & 0.8085 $\pm$ 0.0258          & 0.4475 $\pm$ 0.4094          & 0.7498 $\pm$ 0.0243          \\
\textbf{LAMRec}    & 0.5513 $\pm$ 0.2477          & 0.7771 $\pm$ 0.0460          & 0.5991 $\pm$ 0.1173          \\ \hline
\textbf{\ourmodel} & \textbf{0.9768 $\pm$ 0.0255} & \textbf{0.8186 $\pm$ 0.0146} & \textbf{0.7938 $\pm$ 0.0292} \\ \hline
\end{tabular}
}
\caption{Performance comparison under unseen settings. We report \textit{tF1}, the F1-score specific to the target medicine in each case.}
\end{table}

%% file: tables/appendix_ddi_mimic3.tex
\begin{table}[!ht]
\centering
\caption{Performance comparison for DDI scores evaluated on MIMIC-III.}
\label{tab:ddi3}
\small
\begin{tabular}{c|cccc}
\hline
                   & \multicolumn{4}{c}{\textbf{MIMIC III}}                                                                                                                                                                                                                                                \\ \cline{2-5} 
                   & \textbf{DDI $\downarrow$}                                          & \textbf{wDDI $\downarrow$}                                         & \textbf{aDDI $\downarrow$}                                           & \textbf{awDDI $\downarrow$}                                          \\ \hline
LR                 & \begin{tabular}[c]{@{}c@{}}0.0750\footnotesize $\pm$0.0021\end{tabular}          & \begin{tabular}[c]{@{}c@{}}0.0359\footnotesize $\pm$0.0008\end{tabular}          & \begin{tabular}[c]{@{}c@{}}0.00032\footnotesize $\pm$0.00011\end{tabular}          & \begin{tabular}[c]{@{}c@{}}0.00016\footnotesize $\pm$0.00006\end{tabular}          \\
ECC                & \begin{tabular}[c]{@{}c@{}}0.0726\footnotesize $\pm$0.0019\end{tabular}          & \begin{tabular}[c]{@{}c@{}}0.0349\footnotesize $\pm$0.0007\end{tabular}          & \begin{tabular}[c]{@{}c@{}}0.00030\footnotesize $\pm$0.00014\end{tabular}          & \begin{tabular}[c]{@{}c@{}}0.00015\footnotesize $\pm$0.00008\end{tabular}          \\
Retain             & \begin{tabular}[c]{@{}c@{}}0.0808\footnotesize $\pm$0.0031\end{tabular}          & \begin{tabular}[c]{@{}c@{}}0.0376\footnotesize $\pm$0.0010\end{tabular}          & \begin{tabular}[c]{@{}c@{}}0.00024\footnotesize $\pm$0.00023\end{tabular}          & \begin{tabular}[c]{@{}c@{}}0.00012\footnotesize $\pm$0.00012\end{tabular}          \\
Leap               & \begin{tabular}[c]{@{}c@{}}0.0862\footnotesize $\pm$0.0040\end{tabular}          & \begin{tabular}[c]{@{}c@{}}0.0386\footnotesize $\pm$0.0016\end{tabular}          & \begin{tabular}[c]{@{}c@{}}0.00027\footnotesize $\pm$0.00013\end{tabular}          & \begin{tabular}[c]{@{}c@{}}0.00012\footnotesize $\pm$0.00006\end{tabular}          \\
GAMENet            & \begin{tabular}[c]{@{}c@{}}0.0790\footnotesize $\pm$0.0018\end{tabular}          & \begin{tabular}[c]{@{}c@{}}0.0382\footnotesize $\pm$0.0006\end{tabular}          & \begin{tabular}[c]{@{}c@{}}0.00034\footnotesize $\pm$0.00011\end{tabular}          & \begin{tabular}[c]{@{}c@{}}0.00017\footnotesize $\pm$0.00005\end{tabular}          \\
SafeDrug           & \begin{tabular}[c]{@{}c@{}}0.0624\footnotesize $\pm$0.0008\end{tabular}          & \begin{tabular}[c]{@{}c@{}}0.0301\footnotesize $\pm$0.0002\end{tabular}          & \textbf{\begin{tabular}[c]{@{}c@{}}0.00012\footnotesize $\pm$0.00005\end{tabular}} & \textbf{\begin{tabular}[c]{@{}c@{}}0.00006\footnotesize $\pm$0.00003\end{tabular}} \\
MICRON             & \textbf{\begin{tabular}[c]{@{}c@{}}0.0578\footnotesize $\pm$0.0018\end{tabular}} & \textbf{\begin{tabular}[c]{@{}c@{}}0.0275\footnotesize $\pm$0.0008\end{tabular}} & \begin{tabular}[c]{@{}c@{}}0.00015\footnotesize $\pm$0.00013\end{tabular}          & \begin{tabular}[c]{@{}c@{}}0.00008\footnotesize $\pm$0.00006\end{tabular}          \\
MoleRec            & \begin{tabular}[c]{@{}c@{}}0.0700\footnotesize $\pm$0.0022\end{tabular}          & \begin{tabular}[c]{@{}c@{}}0.0334\footnotesize $\pm$0.0009\end{tabular}          & \begin{tabular}[c]{@{}c@{}}0.00023\footnotesize $\pm$0.00006\end{tabular}          & \begin{tabular}[c]{@{}c@{}}0.00012\footnotesize $\pm$0.00003\end{tabular}          \\
Carmen             & \begin{tabular}[c]{@{}c@{}}0.0875\footnotesize $\pm$0.0029\end{tabular}          & \begin{tabular}[c]{@{}c@{}}0.0409\footnotesize $\pm$0.0005\end{tabular}          & \begin{tabular}[c]{@{}c@{}}0.00026\footnotesize $\pm$0.00009\end{tabular}          & \begin{tabular}[c]{@{}c@{}}0.00012\footnotesize $\pm$0.00004\end{tabular}          \\
LAMRec             & \begin{tabular}[c]{@{}c@{}}0.0621\footnotesize $\pm$0.0031\end{tabular}          & \begin{tabular}[c]{@{}c@{}}0.0294\footnotesize $\pm$0.0017\end{tabular}          & \textbf{\begin{tabular}[c]{@{}c@{}}0.00012\footnotesize $\pm$0.00008\end{tabular}} & \begin{tabular}[c]{@{}c@{}}0.00012\footnotesize $\pm$0.00008\end{tabular}          \\ \hline
OurModel w/o intra & \begin{tabular}[c]{@{}c@{}}0.0831\footnotesize $\pm$0.0022\end{tabular}          & \begin{tabular}[c]{@{}c@{}}0.0392\footnotesize $\pm$0.0012\end{tabular}          & \begin{tabular}[c]{@{}c@{}}0.00016\footnotesize $\pm$0.00004\end{tabular}          & \begin{tabular}[c]{@{}c@{}}0.00008\footnotesize $\pm$0.00002\end{tabular}          \\
OurModel w/o inter & \begin{tabular}[c]{@{}c@{}}0.0833\footnotesize $\pm$0.0017\end{tabular}          & \begin{tabular}[c]{@{}c@{}}0.0395\footnotesize $\pm$0.0008\end{tabular}          & \begin{tabular}[c]{@{}c@{}}0.00020\footnotesize $\pm$0.00006\end{tabular}          & \begin{tabular}[c]{@{}c@{}}0.00011\footnotesize $\pm$0.00003\end{tabular}          \\
OurModel w/o fus   & \begin{tabular}[c]{@{}c@{}}0.0781\footnotesize $\pm$0.0018\end{tabular}          & \begin{tabular}[c]{@{}c@{}}0.0372\footnotesize $\pm$0.0008\end{tabular}          & \begin{tabular}[c]{@{}c@{}}0.00023\footnotesize $\pm$0.00005\end{tabular}          & \begin{tabular}[c]{@{}c@{}}0.00011\footnotesize $\pm$0.00002\end{tabular}          \\ \hline
Ours               & \begin{tabular}[c]{@{}c@{}}0.0782\footnotesize $\pm$0.0026\end{tabular}          & \begin{tabular}[c]{@{}c@{}}0.0373\footnotesize $\pm$0.0012\end{tabular}          & \begin{tabular}[c]{@{}c@{}}0.00020\footnotesize $\pm$0.00008\end{tabular}          & \begin{tabular}[c]{@{}c@{}}0.00010\footnotesize $\pm$0.00004\end{tabular}          \\ \hline
\end{tabular}
\end{table}

%% file: tables/appendix_ddi_mimic4.tex
\begin{table}[!ht]
\centering
\caption{Performance comparison for DDI scores evaluated on MIMIC-IV.}
\label{tab:ddi4}
\small
\begin{tabular}{c|cccc}
\hline
                   & \multicolumn{4}{c}{\textbf{MIMIC IV}}                                                                                                                                                                                                                                                 \\ \cline{2-5} 
                   & \textbf{DDI $\downarrow$}                                          & \textbf{wDDI $\downarrow$}                                         & \textbf{aDDI $\downarrow$}                                           & \textbf{awDDI $\downarrow$}                                          \\ \hline
LR                 & \begin{tabular}[c]{@{}c@{}}0.0877 \footnotesize $\pm$ 0.0015\end{tabular}          & \begin{tabular}[c]{@{}c@{}}0.0418 \footnotesize $\pm$ 0.0004\end{tabular}          & \begin{tabular}[c]{@{}c@{}}0.00010 \footnotesize $\pm$ 0.00005\end{tabular}          & \begin{tabular}[c]{@{}c@{}}0.00005 \footnotesize $\pm$ 0.00003\end{tabular}          \\
ECC                & \begin{tabular}[c]{@{}c@{}}0.0876 \footnotesize $\pm$ 0.0020\end{tabular}          & \begin{tabular}[c]{@{}c@{}}0.0427 \footnotesize $\pm$ 0.0006\end{tabular}          & \begin{tabular}[c]{@{}c@{}}0.00012 \footnotesize $\pm$ 0.00008\end{tabular}          & \begin{tabular}[c]{@{}c@{}}0.00006 \footnotesize $\pm$ 0.00005\end{tabular}          \\
Retain             & \begin{tabular}[c]{@{}c@{}}0.1069 \footnotesize $\pm$ 0.0039\end{tabular}          & \begin{tabular}[c]{@{}c@{}}0.0430 \footnotesize $\pm$ 0.0011\end{tabular}          & \begin{tabular}[c]{@{}c@{}}0.00005 \footnotesize $\pm$ 0.00002\end{tabular}          & \textbf{\begin{tabular}[c]{@{}c@{}}0.00002 \footnotesize $\pm$ 0.00001\end{tabular}} \\
Leap               & \begin{tabular}[c]{@{}c@{}}0.1261 \footnotesize $\pm$ 0.0076\end{tabular}          & \begin{tabular}[c]{@{}c@{}}0.0504 \footnotesize $\pm$ 0.0031\end{tabular}          & \textbf{\begin{tabular}[c]{@{}c@{}}0.00006 \footnotesize $\pm$ 0.00002\end{tabular}} & \begin{tabular}[c]{@{}c@{}}0.00003 \footnotesize $\pm$ 0.00001\end{tabular}          \\
GAMENet            & \begin{tabular}[c]{@{}c@{}}0.0773 \footnotesize $\pm$ 0.0024\end{tabular}          & \begin{tabular}[c]{@{}c@{}}0.0377 \footnotesize $\pm$ 0.0007\end{tabular}          & \begin{tabular}[c]{@{}c@{}}0.00039 \footnotesize $\pm$ 0.00004\end{tabular}          & \begin{tabular}[c]{@{}c@{}}0.00020 \footnotesize $\pm$ 0.00002\end{tabular}          \\
SafeDrug           & \begin{tabular}[c]{@{}c@{}}0.0659 \footnotesize $\pm$ 0.0037\end{tabular}          & \textbf{\begin{tabular}[c]{@{}c@{}}0.0307 \footnotesize $\pm$ 0.0012\end{tabular}} & \textbf{\begin{tabular}[c]{@{}c@{}}0.00006 \footnotesize $\pm$ 0.00002\end{tabular}} & \begin{tabular}[c]{@{}c@{}}0.00003 \footnotesize $\pm$ 0.00001\end{tabular}          \\
MICRON             & \begin{tabular}[c]{@{}c@{}}0.0708 \footnotesize $\pm$ 0.0024\end{tabular}          & \begin{tabular}[c]{@{}c@{}}0.0357 \footnotesize $\pm$ 0.0015\end{tabular}          & \begin{tabular}[c]{@{}c@{}}0.00007 \footnotesize $\pm$ 0.00003\end{tabular}          & \begin{tabular}[c]{@{}c@{}}0.00004 \footnotesize $\pm$ 0.00002\end{tabular}          \\
MoleRec            & \begin{tabular}[c]{@{}c@{}}0.0769 \footnotesize $\pm$ 0.0023\end{tabular}          & \begin{tabular}[c]{@{}c@{}}0.0375 \footnotesize $\pm$ 0.0013\end{tabular}          & \begin{tabular}[c]{@{}c@{}}0.00007 \footnotesize $\pm$ 0.00003\end{tabular}          & \begin{tabular}[c]{@{}c@{}}0.00003 \footnotesize $\pm$ 0.00001\end{tabular}          \\
Carmen             & \begin{tabular}[c]{@{}c@{}}0.0937 \footnotesize $\pm$ 0.0013\end{tabular}          & \begin{tabular}[c]{@{}c@{}}0.0432 \footnotesize $\pm$ 0.0005\end{tabular}          & \textbf{\begin{tabular}[c]{@{}c@{}}0.00006 \footnotesize $\pm$ 0.00002\end{tabular}} & \begin{tabular}[c]{@{}c@{}}0.00003 \footnotesize $\pm$ 0.00001\end{tabular}          \\
LAMRec             & \textbf{\begin{tabular}[c]{@{}c@{}}0.0629 \footnotesize $\pm$ 0.0009\end{tabular}} & \begin{tabular}[c]{@{}c@{}}0.0308 \footnotesize $\pm$ 0.0004\end{tabular}          & \begin{tabular}[c]{@{}c@{}}0.00007 \footnotesize $\pm$ 0.00004\end{tabular}          & \begin{tabular}[c]{@{}c@{}}0.00007 \footnotesize $\pm$ 0.00004\end{tabular}          \\ \hline
OurModel w/o intra & \begin{tabular}[c]{@{}c@{}}0.0924 \footnotesize $\pm$ 0.0022\end{tabular}          & \begin{tabular}[c]{@{}c@{}}0.0438 \footnotesize $\pm$ 0.0012\end{tabular}          & \begin{tabular}[c]{@{}c@{}}0.00010 \footnotesize $\pm$ 0.00004\end{tabular}          & \begin{tabular}[c]{@{}c@{}}0.00005 \footnotesize $\pm$ 0.00002\end{tabular}          \\
OurModel w/o inter & \begin{tabular}[c]{@{}c@{}}0.0922 \footnotesize $\pm$ 0.0017\end{tabular}          & \begin{tabular}[c]{@{}c@{}}0.0438 \footnotesize $\pm$ 0.0008\end{tabular}          & \begin{tabular}[c]{@{}c@{}}0.00009 \footnotesize $\pm$ 0.00006\end{tabular}          & \begin{tabular}[c]{@{}c@{}}0.00005 \footnotesize $\pm$ 0.00003\end{tabular}          \\
OurModel w/o fus   & \begin{tabular}[c]{@{}c@{}}0.0928 \footnotesize $\pm$ 0.0018\end{tabular}          & \begin{tabular}[c]{@{}c@{}}0.0440 \footnotesize $\pm$ 0.0008\end{tabular}          & \begin{tabular}[c]{@{}c@{}}0.00009 \footnotesize $\pm$ 0.00005\end{tabular}          & \begin{tabular}[c]{@{}c@{}}0.00004 \footnotesize $\pm$ 0.00002\end{tabular}          \\ \hline
Ours               & \begin{tabular}[c]{@{}c@{}}0.0924 \footnotesize $\pm$ 0.0026\end{tabular}          & \begin{tabular}[c]{@{}c@{}}0.0439 \footnotesize $\pm$ 0.0012\end{tabular}          & \begin{tabular}[c]{@{}c@{}}0.00009 \footnotesize $\pm$ 0.00008\end{tabular}          & \begin{tabular}[c]{@{}c@{}}0.00004 \footnotesize $\pm$ 0.00004\end{tabular}          \\ \hline
\end{tabular}
\end{table}

%% file: tables/appendix_unseen_R03A.tex
\begin{table}[!ht]
\centering
\label{tab:unseen_r03a}
\begin{tabular}{l|rrrrrrr}
\hline
\textbf{R03A}      & \multicolumn{1}{l}{\textbf{Jaccard} $\uparrow$} & \multicolumn{1}{l}{\textbf{PRAUC} $\uparrow$} & \multicolumn{1}{l}{\textbf{F1} $\uparrow$} & \multicolumn{1}{l}{\textbf{med}} & \multicolumn{1}{l}{\textbf{tF1} $\uparrow$} & \multicolumn{1}{l}{\textbf{tPrec} $\uparrow$} & \multicolumn{1}{l}{\textbf{tRecall} $\uparrow$} \\ \hline
\textbf{GAMENet}   & 0.4027                               & 0.4689                             & 0.5468                          & 46.31                            & 0.6283                           & 0.6509                             & 0.6225                               \\
\textbf{SafeDrug}  & 0.4269                               & 0.7096                             & 0.5873                          & 24.75                            & 0.5766                           & 0.5944                             & 0.5729                               \\
\textbf{MoleRec}   & 0.4435                               & \textbf{0.7118}                    & 0.6029                          & 25.10                            & 0.6635                           & 0.6679                             & 0.6671                               \\
\textbf{Carmen}    & 0.4362                               & 0.7065                             & 0.5956                          & 24.98                            & 0.8085                           & 0.8139                             & 0.8074                               \\
\textbf{LAMRec}    & 0.4260                               & 0.6847                             & 0.5842                          & 27.62                            & 0.5513                           & 0.8659                             & 0.4465                               \\ \hline
\textbf{Our Model} & \textbf{0.4619}                      & 0.7017                             & \textbf{0.6234}                 & 28.8741                          & \textbf{0.9768}                  & \textbf{1.0000}                    & \textbf{0.9556}                      \\ \hline
\end{tabular}
\vspace{2em}

\begin{tabular}{l|rrrrrrr}
\hline
\textbf{N03A}      & \multicolumn{1}{l}{\textbf{Jaccard} $\uparrow$} & \multicolumn{1}{l}{\textbf{PRAUC} $\uparrow$} & \multicolumn{1}{l}{\textbf{F1} $\uparrow$} & \multicolumn{1}{l}{\textbf{med}} & \multicolumn{1}{l}{\textbf{tF1} $\uparrow$} & \multicolumn{1}{l}{\textbf{tPrec} $\uparrow$} & \multicolumn{1}{l}{\textbf{tRecall} $\uparrow$} \\ \hline
\textbf{GAMENet}   & 0.4237                               & 0.6294                             & 0.5721                          & 37.87                            & 0.5279                           & 0.5344                             & 0.5295                               \\
\textbf{SafeDrug}  & 0.4384                               & 0.7249                             & 0.5989                          & 23.36                            & 0.4440                           & 0.4472                             & 0.4464                               \\
\textbf{MoleRec}   & 0.4555                               & \textbf{0.7181}                    & 0.6152                          & 24.02                            & 0.5109                           & 0.5121                             & 0.5166                               \\
\textbf{Carmen}    & 0.4476                               & 0.7051                             & 0.6063                          & 26.08                            & 0.4475                           & 0.4531                             & 0.4442                               \\
\textbf{LAMRec}    & \textbf{0.5306}                      & \textbf{0.7790}                    & \textbf{0.6769}                 & 26.92                            & 0.7771                           & 0.8897                             & 0.6959                               \\ \hline
\textbf{Our Model} & \textbf{0.4617}                      & 0.7338                             & \textbf{0.6193}                 & 23.30                            & \textbf{0.8186}                  & \textbf{0.8403}                    & \textbf{0.7991}                      \\ \hline
\end{tabular}

\vspace{2em}
\begin{tabular}{l|rrrrrrr}
\hline
\textbf{C01C}      & \multicolumn{1}{l}{\textbf{Jaccard} $\uparrow$} & \multicolumn{1}{l}{\textbf{PRAUC} $\uparrow$} & \multicolumn{1}{l}{\textbf{F1} $\uparrow$} & \multicolumn{1}{l}{\textbf{med}} & \multicolumn{1}{l}{\textbf{tF1} $\uparrow$} & \multicolumn{1}{l}{\textbf{tPrec} $\uparrow$} & \multicolumn{1}{l}{\textbf{tRecall} $\uparrow$} \\ \hline
\textbf{GAMENet}   & 0.4684                               & 0.6170                             & 0.6167                          & 45.72                            & 0.6225                           & 0.6320                             & 0.6219                               \\
\textbf{SafeDrug}  & 0.4693                               & 0.7463                             & 0.6310                          & 27.67                            & 0.4709                           & 0.4826                             & 0.4677                               \\
\textbf{MoleRec}   & 0.4802                               & \textbf{0.7426}                    & 0.6400                          & 28.29                            & 0.5094                           & 0.5212                             & 0.5065                               \\
\textbf{Carmen}    & 0.4856                               & 0.7356                             & 0.6439                          & 30.85                            & 0.7498                           & 0.8731                             & 0.6602                               \\
\textbf{LAMRec}    & \textbf{0.4728}                      & \textbf{0.7257}                    & \textbf{0.6295}                 & 31.48                            & 0.5991                           & \textbf{0.8957}                    & 0.4622                               \\ \hline
\textbf{Our Model} & \textbf{0.4917}                      & \textbf{0.7491}                    & \textbf{0.6498}                 & 31.79                            & \textbf{0.7938}                  & \textbf{0.8669}                    & \textbf{0.7338}                      \\ \hline

\end{tabular}
\caption{Performance comparison under unseen settings for target medication R03A, N03A, and C01C.}
\end{table}

%% file: tables/appendix_graph_analysis.tex
\begin{table}[!ht]
\centering
\textbf{Retained Strong-Prior Edges} \\
\textbf{Recommended Drugs:} \textit{PARASYMPATHOMIMETICS}, \textit{DRUGS FOR FUNCTIONAL GASTROINTESTINAL DISORDERS}
\vspace{0.3em}
\resizebox{0.9\textwidth}{!}{%
\begin{tabular}{ll}
\toprule
\textbf{Source} & \textbf{Target} \\
\midrule
Septic arterial embolism & OTHER ANTIBACTERIALS in ATC \\
Septic arterial embolism & OPIOID ANALGESICS \\
Septic arterial embolism & OTHER ANALGESICS AND ANTIPYRETICS in ATC \\
DRUGS FOR PEPTIC ULCER AND GORD & DRUGS FOR FUNCTIONAL GASTROINTESTINAL DISORDERS \\
DRUGS FOR CONSTIPATION & DRUGS FOR FUNCTIONAL GASTROINTESTINAL DISORDERS \\
DRUGS FOR FUNCTIONAL GASTROINTESTINAL DISORDERS & ANESTHETICS, GENERAL \\
PROPULSIVES & DRUGS FOR PEPTIC ULCER AND GORD \\
ANTIMYCOTICS FOR SYSTEMIC USE & DRUGS FOR PEPTIC ULCER AND GORD \\
ANTITHROMBOTIC AGENTS & ANTIMYCOTICS FOR SYSTEMIC USE \\
PARASYMPATHOMIMETICS & DRUGS FOR PEPTIC ULCER AND GORD \\
PARASYMPATHOMIMETICS & DRUGS FOR CONSTIPATION \\
\bottomrule
\end{tabular}%
}
\vspace{0.3em}

\textbf{Retained Weak-Prior Edges} \\
\textbf{Recommended Drugs:} \textit{INTESTINAL ANTIINFECTIVES}

\vspace{0.3em}
\resizebox{0.9\textwidth}{!}{%
\begin{tabular}{ll}
\toprule
\textbf{Source} & \textbf{Target} \\
\midrule
DRUGS FOR CONSTIPATION & POSTERIOR PITUITARY LOBE HORMONES \\
DRUGS FOR CONSTIPATION & VITAMIN K AND OTHER HEMOSTATICS \\
HYPOTHALAMIC HORMONES & DRUGS FOR CONSTIPATION \\
LOCAL ANESTHETICS & DRUGS FOR CONSTIPATION \\
ANTICHOLINERGIC AGENTS & DRUGS FOR CONSTIPATION \\
ANTITHROMBOTIC AGENTS & DRUGS FOR CONSTIPATION \\
ANTITHROMBOTIC AGENTS & BILE THERAPY DRUGS \\
ANTITHROMBOTIC AGENTS & POSTERIOR PITUITARY LOBE HORMONES \\
ANTITHROMBOTIC AGENTS & HYPOTHALAMIC HORMONES \\
ANTIPRURITICS & DRUGS FOR CONSTIPATION \\
ANTIEPILEPTICS & BILE THERAPY DRUGS \\
HYPOTHALAMIC HORMONES & MUSCLE RELAXANTS, PERIPHERALLY ACTING AGENTS \\
LOCAL ANESTHETICS & OTHER ANTIBACTERIALS in ATC \\
ANTICHOLINERGIC AGENTS & OTHER ANTIBACTERIALS in ATC \\
\bottomrule
\end{tabular}%
}
\vspace{0.3em}

\textbf{Pruned Strong-Prior Edges} \\
\vspace{0.3em}
\resizebox{0.9\textwidth}{!}{%
\begin{tabular}{ll}
\toprule
\textbf{Disease} & \textbf{Drug} \\
\midrule
Intermediate coronary syndrome & DRUGS FOR CONSTIPATION \\
Intermediate coronary syndrome & OTHER ANALGESICS AND ANTIPYRETICS in ATC \\
Pleurisy (TB excluded) & DRUGS FOR PEPTIC ULCER AND GORD \\
Pleurisy (TB excluded) & DRUGS FOR CONSTIPATION \\
Pleurisy (TB excluded) & OTHER MINERAL SUPPLEMENTS in ATC \\
Pleurisy (TB excluded) & ANTITHROMBOTIC AGENTS \\
Pleurisy (TB excluded) & OPIOID ANALGESICS \\
Pleurisy (TB excluded) & OTHER ANALGESICS AND ANTIPYRETICS in ATC \\
Retinal hemorrhage & DRUGS FOR PEPTIC ULCER AND GORD \\
Retinal hemorrhage & ANTIEMETICS AND ANTINAUSEANTS \\
Retinal hemorrhage & DRUGS FOR CONSTIPATION \\
Retinal hemorrhage & POTASSIUM SUPPLEMENTS \\
Retinal hemorrhage & OTHER MINERAL SUPPLEMENTS in ATC \\
Retinal hemorrhage & ANTITHROMBOTIC AGENTS \\
Retinal hemorrhage & HIGH-CEILING DIURETICS \\
Retinal hemorrhage & CORTICOSTEROIDS FOR SYSTEMIC USE, PLAIN \\
Retinal hemorrhage & OTHER BETA-LACTAM ANTIBACTERIALS in ATC \\
\bottomrule
\end{tabular}%
}
\end{table}